# Modeling Ambiguity in a Multi-Agent System

Christof Monz

Institute for Logic, Language and Computation (ILLC)
University of Amsterdam, Plantage Muidergracht 24
1018 TV Amsterdam, The Netherlands. E-mail: `christof@science.uva.nl`

**Abstract**

This paper investigates the formal pragmatics of ambiguous expressions by modeling ambiguity in a multi-agent system. Such a framework allows us to give a more refined notion of the kind of information that is conveyed by ambiguous expressions. We analyze how ambiguity affects the knowledge of the dialog participants and, especially, what they know about each other after an ambiguous sentence has been uttered. The agents communicate with each other by means of a `tell`-function, whose application is constrained by an implementation of some of Grice's maxims. The information states of the multi-agent system itself are represented as a Kripke structures and `tell` is an update function on those structures. This framework enables us to distinguish between the information conveyed by ambiguous sentences vs. the information conveyed by disjunctions, and between semantic ambiguity vs. perceived ambiguity.

## 1 Introduction

The ambiguity of natural language poses problems for any formal theory within Linguistics, Philosophy, Cognitive Psychology or Artificial Intelligence. If one tries to set up a formal framework dealing with ambiguous expressions, one immediately faces interesting fundamental questions about the nature of ambiguity: How should the formal semantics of an ambiguous expression be defined? What is the information conveyed by ambiguous expressions? When is an expression perceived as ambiguous? The last question is relevant since speakers are often not aware of the fact that they have said something which is ambiguous.

These questions show that ambiguity mainly becomes relevant in situations where more than one person is involved. In this paper, we analyze when ambiguities arise by looking at a multi-agent system, where the agents also communicate with ambiguous statements. Analyzing ambiguity in a multi-agent system is promising since it allows us to create an artificial dialog situation where the environment is clearly defined and the issues mentioned above can be analyzed under sterile conditions.

Additionally, multi-agent systems are not only interesting as a tool for semanticists to formalize some pragmatic aspects. Existing multi-agent communication languages like KQML (Labrou and Finin 1994) and ACL (FIPA 1999) use some of Grice's *cooperative principles* (Grice 1989) to define the semantics of the agent language and make use of pragmatic notions stemming from speech act theory (cf. Searle 1969) to implement more complex forms of agent communication.

## 2 A Multi-Agent System with Ambiguity

The framework as it is proposed here, is to some extent similar to the contextual approach of Buvač (1996), who also uses a modal logic to formalize ambiguity. On the other hand, his approach is restricted to a single-agent scenario, and therefore not appropriate to answer the questions we mentioned in the preceding section.

### 2.1 The Framework

The scenario is very basic. Each member of a group of $n$ agents knows some facts about the world and the other agents. The only action taking place, is that one of the agents tells a group of other agents what he knows. Is is assumed that he only tells things that he knows to be true and that the other agents do not doubt what he is saying. Two agents might differ in what they know, but for simplicity, it is assumed that their knowledge is non-conflicting.

To model the information state of the agents, we use valued rooted Kripke structures $(\mathfrak{M}, w, V)$: $\mathfrak{M} = \langle W, \{R_i\}_{i \in \mathcal{A}}, \mathcal{A} \rangle$, $w \in W$, and $V$ is a valuation function, $V : W \to \mathbb{P}^2$, $\mathbb{P}^2$ being the powerset of the set of propositional variables $\mathbb{P}$. As usual, $W$ is a set of possible worlds, $\mathcal{A}$ a finite set of Agents, and $\{R_i\}_{i \in \mathcal{A}}$ is a set of accessibility relations between worlds for each agent $i \in \mathcal{A}$. We say that $wR_iw'$ if agent $i$ considers world $w'$ possible in world $w$. Knowledge is expressed by a set of modal operators $\{K_i\}_{i \in \mathcal{A}}$, where $(\mathfrak{M}, w) \models K_i\varphi$ iff $(\mathfrak{M}, w') \models \varphi$ for all $w'$ such that $wR_iw'$. In the sequel, we sometimes omit the valuation: $(\mathfrak{M}, w) \models \varphi$ iff $(\mathfrak{M}, w, V) \models \varphi$, for arbitrary $V$.

Group knowledge ($E$) and common knowledge ($C$) are defined as in Fagin et al. (1995): $(\mathfrak{M}, w) \models E_G\varphi$ iff $(\mathfrak{M}, w) \models \bigwedge_{i \in G} K_i\varphi$ and $(\mathfrak{M}, w) \models C_G\varphi$ iff $(\mathfrak{M}, w) \models E_G(\varphi \wedge C_G\varphi)$. Information structures are fully introspective and serial; i.e., they are in $KD45_n$: For all $w \in W$ and $i \in \mathcal{A}$ it holds that

(D)  $(\mathfrak{M}, w) \models \neg K_i\bot$ (seriality)
(4)  $(\mathfrak{M}, w) \models K_i\varphi \to K_iK_i\varphi$ (positive introspection)
(5)  $(\mathfrak{M}, w) \models \neg K_i\varphi \to K_i\neg K_i\varphi$ (negative introspection)

The agents are endowed with very limited communicative capabilities. In fact, the only communicative acts are of the form `tell`$(i, G, S)$, which is defined as follows:

**Definition 1 (`tell`)** `tell` is a function from valued rooted Kripke structures to valued rooted Kripke structures. It can be seen as an update function. Additionally, `tell` has three parameters, $(i, \{G\}, S)$, where $i$ is an agent (the speaker), $G \subseteq \mathcal{A}\backslash\{i\}$ is a group of agents (the hearers),[1] and $S$ is a natural language sentence belonging to some (not further specified) fragment $\mathcal{L}^{eng}$ of English.

---
[1] The restriction that the speaker is not part of the group of hearers is mainly due to formal convenience, as it simplifies the implementation of some of the cooperative principles.

Although being desirable, it is not possible in the proposed framework that two communicative acts take place at the same time. Allowing two or more `tell`-actions to take place at the same time, would require parallel updating. Again, for simplicity, we do not consider this possibility.

A function $\tau$ from $\mathcal{L}^{eng}$ to $\mathcal{L}^{PL0}$ (propositional logic) generates the possible semantic representations of $S$. Actually, $\tau(S)$ returns the set of equivalence classes of the readings, such that readings which are equivalent belong to the same class. $[\varphi]$ indicates the equivalence class $\varphi$ belongs to.

**Definition 2 (Semantic Ambiguity)** If $S \in \mathcal{L}^{eng}$, then $S$ is said to be semantically ambiguous iff $|\tau(S)| > 1$, i.e., there are at least two non-equivalent ways to represent the semantics of $S$.

The fact that $s$ is semantically ambiguous does not mean that it is also *perceived* as ambiguous by the speaker or the hearer(s), cf. Poesio (1996). In Section 2.2, we will see how to define the difference between semantic and perceived ambiguity formally.

We are aware of the fact that propositional logic is not expressive enough to be an appropriate representation language for the semantics of natural language utterances, but this restriction allows us to blend out further intricacies that can arise in a multi-agent system for natural language dialogs (see, e.g., Francez and Berg 1994), so we can focus on issues that are strictly related to the problem of ambiguity.

## 2.2 Implementing Cooperative Principles

Before we discuss in more detail how ambiguity is treated, we pose some general constraints on the execution of communicative acts. These constraints are based on *Grice's maxims*, cf. Grice (1989), and can be considered as a partial implementation of Grice's maxims in a multi-agent system, see also Labrou and Finin (1994) and FIPA (1999) for different ways of integrating Grice's maxims into an agent communication language.

**Definition 3 (Cooperative Principles)** Given a valued rooted Kripke structure of the form $(\mathfrak{M}, w, V)$ representing the current information state, the following constraints are imposed on the application of `tell`. If $\texttt{tell}(i, G, S)(\mathfrak{M}, w, V) = (\mathfrak{M}', w', V')$, then $\exists [\varphi] \in \tau(S)$ such that

(i) $(\mathfrak{M}, w, V) \models K_i \varphi$
   **Maxim of Quality**: *do not say that for which you lack adequate evidence.* This is implemented by requiring that a speaker knows that at least one of the readings of $S$ is true.

(ii) $(\mathfrak{M}, w, V) \models K_i \neg C_{\{i\} \cup G} \varphi$
   **Maxim of Quantity**: *make your contribution as informative as is required for the current purposes of the exchange.* Agent $i$ knows that at least one reading of $S$ is not part of the common knowledge.

(iii) $\forall [\psi] \in \tau(S) \setminus \{[\varphi]\} (\mathfrak{M}, w, V) \models K_i(C_{\{i\} \cup G} \psi \vee C_{\{i\} \cup G} \neg \psi)$
   **Maxim of Manner**: *avoid ambiguity.* Agent *i* knows that it is common knowledge amongst the speaker and the hearers that either all readings $\psi$ of *S*, such that $[\psi] \neq [\varphi]$, are true (and therefore uninformative) or false (an therefore conflicting).

(iv) $(\mathfrak{M}', w', V') \models C_{\{i\} \cup G} \varphi$
   **Grounding Criterion**: After telling *S*, $\varphi$ is common knowledge of the hearers and the speaker (*i*).

Similar to Labrou and Finin (1994), these conditions are divided into pre-conditions that have to hold before a communicative action can be executed (i.e., before updating the original Kripke model) and post-conditions that describe what has to hold afterwards.

(i)–(iii) are the pre-conditions of applying `tell` to a valued Kripke structure. (iv) is the post-condition, where the speaker's contribution is added to the common knowledge, see e.g. Clark and Schaefer (1992).

Updating in a multi-agent system where the pre-conditions (i)–(iii) are implemented as hard constraints will always yield a resulting state which satisfies the post-condition (iv). I.e., although the agents can communicate with semantically ambiguous sentences, it cannot happen that such a sentence is also perceived as ambiguous by the hearer(s).

On the other hand, if the pre-conditions (i)–(iii) are implemented as default constraints, and, for instance, the *Maxim of Manner* is violated by a speaker, the *Grounding Criterion* is not guaranteed to hold.

**Definition 4 (Perceived Ambiguity)** Let *S* be a semantically ambiguous sentence, as defined in Definition 2, $\mathtt{tell}(i, G, S)(\mathfrak{M}, w, V) = (\mathfrak{M}', w', V')$, and the following conditions hold:

1. $(\mathfrak{M}', w', V') \models \neg C_{\{i\} \cup G} \varphi$
2. $(\mathfrak{M}', w', V') \models E_{\{i\} \cup G}(\bigvee_{[\varphi] \in \tau(S)} C_{\{i\} \cup G} \varphi)$

Then, we say that *S* is perceived as ambiguous.

Detection of ambiguity is also important for building intelligent dialog systems, because it can inform the system to apply repair strategies, cf. McRoy and Hirst (1995).

This illustrates also the difference between ambiguity and disjunction. If an ambiguous sentence *S* is treated the same way as the disjunction of its *m* readings $\bigvee_{k=1}^{m} \varphi_k$, it would result in a weaker post-condition of the form $(\mathfrak{M}', w', V') \models C_{\{i\} \cup G} \bigvee_{k=1}^{m} \varphi_k$.

Finally, we mention another important constraint on updating, namely the preservation of known facts.

**Definition 5 (Information Increase)** Let $\{K_i\}_{i \in \mathcal{A}}^*$ be the set of finite concatenations of elements of $\{K_i\}_{i \in \mathcal{A}}$, including the empty sequence $\varepsilon$. If $\vec{K} \in \{K_i\}_{i \in \mathcal{A}}^*$ and $(\mathfrak{M}, w, V) \models \vec{K}\varphi$, then $\mathtt{tell}(i, G, S)(\mathfrak{M}, w, V) \models \vec{K}\varphi$.

This definition of information increase, which is a more general reformulation of Groeneveld's *descriptive information increase*, cf. Groeneveld (1995).

## 2.3 Updating with Ambiguous Information

Let us consider an example. The Kripke-structures depicted in Figure 1 and Figure 2, represent the information state of Agent 1 and Agent 2, respectively.[2] We restrict ourselves to two propositional variables $p$ and $q$. Agent 1 does know that $p$ holds, but is uncertain about the truth of $q$. In addition, he does not know whether Agent 2 knows $p$ or $q$, or whether Agent 2 knows whether Agent 1 knows $p$ or $q$, etc.

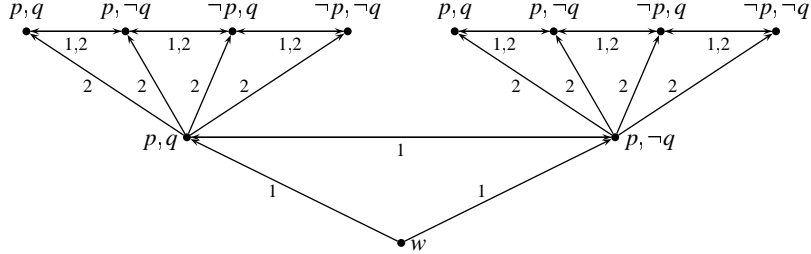

Figure 1: A Kripke model representing the information state of Agent 1

More formally, the following holds:

1. $(\mathfrak{M}, w, V) \models K_1 p \land \neg K_1 q$
2. $(\mathfrak{M}, w, V) \models \neg K_1 K_2 (C_{\{1,2\}} p \lor C_{\{1,2\}} q)$

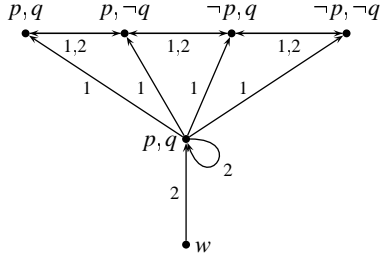

Figure 2: A Kripke model representing the information state of Agent 2

Agent 2, on the other hand does know:

1. $(\mathfrak{M}, w, V) \models K_2 (p \land q)$
2. $(\mathfrak{M}, w, V) \models \neg K_2 (K_1 p \lor K_1 q)$

If Agent 1 tells Agent 2 that $S$ is the case, where $\tau(S) = \{p, q\}$, then the model in Figure 1 and Figure 2 has to be updated with $\texttt{tell}(1, \{2\}, S)$. Again, focusing on the information state of Agent 1, this results in the Kripke-structures displayed by Figure 3 and Figure 4.

---
[2] Note that both figures belong to the same Kripke structure. The split-up is entirely due to space limitations.

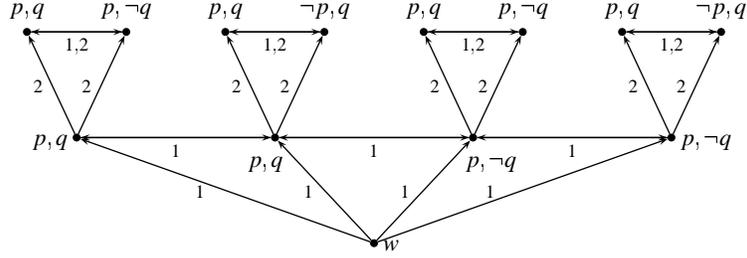

Figure 3: The information state of Agent 1 after updating with *S*

Having told *S* to Agent 2, Agent 1 knows that Agent 2 either knows that *p* holds or that he knows that *q* holds, i.e., $\texttt{tell}(1,\{2\},S)(\mathfrak{M},w,V) \models K_1(K_2 p \vee K_2 q)$.

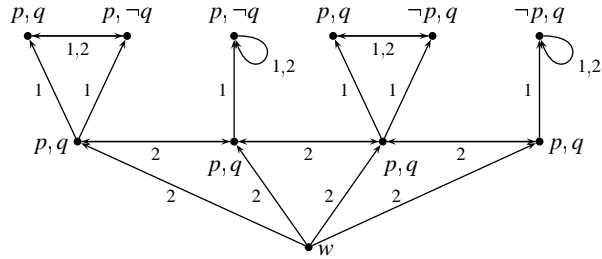

Figure 4: The information state of Agent 2 after updating with *S*

Determining the resulting information state of Agent 2 is a bit more complex. First of all, to which extent should Agent 2 obey the cooperative principles? We know that Agent 1 violated the maxim of manner, but Agent 1 does not know that. For simplicity, we assume that the hearer(s) give the speaker full credit, which means that they think that the speaker obeys all cooperative principles.

Turning to Figure 4, there are four possibilities, two for each reading of *S*. Either Agent 2 thinks that Agent 1 intended to say *p*, because he is uncertain about the truth of *q* or because he knows that *q* is not true; and analogously for *q*, the other reading of *S*.

## 3 Conclusions

We have seen that, to some extent, ambiguity can be modeled in a multi-agent system. It became also clear that this is certainly a non-trivial task, involving a lot of intricacies, most of which we have simplified in this paper. The presented framework is mainly intended to give a rough idea how to formalize ambiguity in a multi-agent system, where several extensions are still needed. In particular, we intend to put future efforts into tackling two problems. How to extend the framework to first-order (dynamic) logic, and which other pragmatic principles can be implemented?

**Acknowledgments.** The author was supported by the Physical Sciences Council with financial support from the Netherlands Organization for Scientific Research (NWO), project 612-13-001.